\documentclass[letterpaper, 10 pt, conference]{ieeeconf}

\IEEEoverridecommandlockouts

\title{\LARGE \bf
    CATNAV: Cached Vision-Language Traversability for Efficient Zero-Shot Robot Navigation}

\author{Aditya Potnis, Francisco Affonso, Shreya Gummadi,
Naveen Kumar Uppalapati, Girish Chowdhary
\thanks{All authors are with the Field Robotics Engineering and Science Hub (FRESH), Illinois Autonomous Farm, University of Illinois at Urbana-Champaign (UIUC), IL}
\thanks{Correspondence to \texttt{\{apotnis2, girishc\}@illinois.edu}}
}
\usepackage{booktabs}
\usepackage{multirow}
\usepackage{makecell}
\usepackage{colortbl}
\usepackage{xcolor}
\usepackage{pifont}

\usepackage{graphicx}
\usepackage{amsfonts}
\usepackage{amsmath}
\usepackage{graphicx}
\usepackage{makecell}
\usepackage{algorithm}
\usepackage{algpseudocode}
\usepackage{stfloats}
\usepackage{xcolor}
\usepackage{pgf}

\usepackage[most]{tcolorbox}

\definecolor{green_traj}{RGB}{56,255,20}
\definecolor{blue_traj}{RGB}{13, 8, 135}

\begin{document}

\maketitle

\thispagestyle{empty}
\pagestyle{empty}

\begin{abstract}
Navigating unstructured environments requires assessing traversal risk relative to a robot's physical capabilities, a challenge that varies across embodiments. We present CATNAV, a cost-aware traversability navigation framework that leverages multimodal LLMs for zero-shot, embodiment-aware costmap generation without task-specific training. We introduce a visuosemantic caching mechanism that detects scene novelty and reuses prior risk assessments for semantically similar frames, reducing online VLM queries by 85.7\%. Furthermore, we introduce a VLM-based trajectory selection module that evaluates proposals through visual reasoning to choose the safest path given behavioral constraints. We evaluate CATNAV on a quadruped robot across indoor and outdoor unstructured environments, comparing against state-of-the-art vision-language-action baselines. Across five navigation tasks, CATNAV achieves 10 percentage point higher average goal-reaching rate and 33\% fewer behavioral constraint violations.
\vspace{-15pt}
\end{abstract}
\section{Introduction}
Deploying robots in unstructured environments such as farms, sidewalks, and construction sites requires reasoning not just about \textit{what} is in the scene, but about the consequences of interacting with it. A muddy patch may be traversable for a tracked vehicle but treacherous for a legged robot; a child near a crosswalk demands different caution than a traffic cone in the same location. This kind of semantic consequence modeling remains a central challenge in robot navigation, even as stronger edge compute and larger training datasets have dramatically expanded the scope of deployable systems.

Often, navigation depends on the environment and the context in which the robot operates. Outdoor driving requires a clear understanding of what is traversable (roads, bumps, mud, grass, etc.) and which dynamic obstacles (e.g., cars, cyclists, etc.) are safe or unsafe in proximity. By contrast, indoor navigation demands socially compliant behaviors, with greater care around people, fragile objects, and indoor risks such as spills or soft carpets. Moreover, indoor tasks generally require semantically specified goals or prior maps, whereas outdoor goals are more easily expressed with GNSS-assisted waypoints due to better accessibility, landmark sparsity, and lower absolute accuracy requirements.

Learned traversability methods~\cite{gasparino2024wayfaster,wrizz2024,schreiber_you_2025,frey23fast,elnoor_vi-lad_2025} achieve strong performance but require extensive, diverse training data and struggle to generalize under out-of-distribution conditions without retraining.

Multi-modal large language models offer a promising alternative: their zero-shot \textit{semantic consequence reasoning}, e.g., identifying that a puddle may be slippery or that tall grass may conceal obstacles~\cite{google2025gemini2.5}, combined with advances in open-vocabulary segmentation~\cite{luddecke2022image} and image-text alignment~\cite{radford2021learning}, has inspired hybrid VLM-assisted navigation. However, existing approaches face a fundamental trade-off. Costmap-based methods~\cite{du_vl-nav_2025,yokoyama2024vlfm,weerakoon2025behav} are compute-efficient but reduce VLM knowledge to scalar costs, discarding the rich consequence reasoning that makes VLMs powerful. Trajectory-level and MPC-based methods~\cite{sathyamoorthy2024convoi,song2024vlm,martinez2025hey} better preserve this reasoning but require frequent online queries, inducing latency and cost even when the semantic context has not meaningfully changed.

Vision-language-action (VLA) models~\cite{castro_vamos_2025,hirose_omnivla_2025} learn navigation policies directly from multi-modal data, but these end-to-end approaches implicitly encode navigational preferences in network weights rather than explicitly reasoning about semantic consequences, limiting their interpretability and adaptability to novel scenarios without retraining.

\begin{table*}[t]
\centering
\caption{Comparison of navigation methods against state-of-the-art.
\textcolor{green!60!black}{\ding{51}} = fully supported,
$\sim$ = partial/limited,
\textcolor{gray}{--} = not addressed.}
\label{tab:comparison}
\renewcommand{\arraystretch}{1.25}
\setlength{\tabcolsep}{5pt}
\resizebox{\textwidth}{!}{
\begin{tabular}{c | cc | cc | ccc}
\toprule
\multirow{2}{*}{\centering \textbf{Method}}
  & \multicolumn{2}{c|}{\textbf{Traversability}}
  & \multicolumn{2}{c|}{\textbf{Planning}}
  & \multicolumn{3}{c}{\textbf{VLM Efficiency}} \\
\cmidrule(lr){2-3}\cmidrule(lr){4-5}\cmidrule(l){6-8}
  & \makecell{Zero-Shot \\ Travers.}
  & \makecell{Dynamic Obj. \\ Segmentation}
  & \makecell{Motion Prim. \\ Aware Plan.}
  & \makecell{Multi-Embod. \\ Support}
  & \makecell{Multi-Modal \\ Costmap}
  & \makecell{Novelty-Driven \\ VLM Polling}
  & \makecell{Low-Compute \\ VLM Use} \\
\midrule
WayFASTER / W-RIZZ / CHUNGUS~\cite{gasparino2024wayfaster,wrizz2024,schreiber_you_2025}
  & \textcolor{gray}{--} & \textcolor{green!60!black}{\ding{51}}
  & \textcolor{gray}{--} & \textcolor{gray}{--}
  & \textcolor{gray}{--} & $\sim$
  & \textcolor{green!60!black}{\ding{51}} \\

BEHAV~\cite{weerakoon2025behav}
  & $\sim$ & \textcolor{gray}{--}
  & \textcolor{gray}{--} & \textcolor{gray}{--}
  & \textcolor{green!60!black}{\ding{51}} & \textcolor{gray}{--}
  & \textcolor{green!60!black}{\ding{51}} \\

ConVOI~\cite{sathyamoorthy2024convoi}
  & $\sim$ & \textcolor{gray}{--}
  & \textcolor{gray}{--} & \textcolor{gray}{--}
  & \textcolor{gray}{--} & \textcolor{green!60!black}{\ding{51}}
  & \textcolor{gray}{--} \\

VLM-Social-Nav / Hey Robot~\cite{song2024vlm,martinez2025hey}
  & $\sim$ & \textcolor{gray}{--}
  & \textcolor{gray}{--} & \textcolor{gray}{--}
  & \textcolor{gray}{--} & \textcolor{gray}{--}
  & \textcolor{gray}{--} \\

VAMOS / OmniVLA~\cite{castro_vamos_2025,hirose_omnivla_2025}
  & $\sim$ & \textcolor{gray}{--}
  & \textcolor{gray}{--} & \textcolor{green!60!black}{\ding{51}}
  &  $\sim$ & \textcolor{gray}{--}
  & \textcolor{gray}{--} \\

\midrule
\textbf{Ours (CATNAV)}
  & \textcolor{green!60!black}{\ding{51}} & \textcolor{green!60!black}{\ding{51}}
  & \textcolor{green!60!black}{\ding{51}} & \textcolor{green!60!black}{\ding{51}}
  & \textcolor{green!60!black}{\ding{51}} & \textcolor{green!60!black}{\ding{51}}
  & \textcolor{green!60!black}{\ding{51}} \\
\bottomrule
\end{tabular}
}
\end{table*}

Vision-language models (VLMs), trained on internet-scale data pairing images with natural language, offer a promising path toward semantic consequence modeling: their broad world knowledge enables zero-shot reasoning about the consequences of traversing or approaching novel objects without task-specific training data.

\begin{figure}[t]
\centering
\includegraphics[width=\columnwidth]{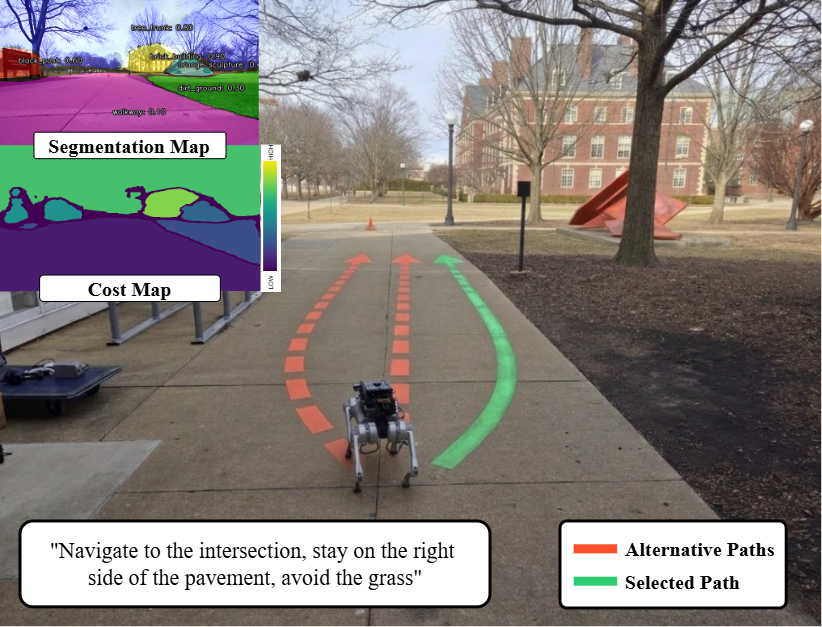}
\caption{We present CATNAV, a zero-shot cost and embodiment aware traversability navigation method that is able to navigate in unknown environments without fine-tuning.}
\label{fig:teaser}
\vspace{-18pt}
\end{figure}
CATNAV addresses this gap by leveraging VLM semantic consequence reasoning for costmap generation while minimizing redundant inference through novelty-driven polling, querying the VLM only when the scene presents semantically novel situations that warrant updated risk assessment. We make the following contributions:
\begin{enumerate}
    \item A \textbf{zero-shot, embodiment-aware costmap generation framework} that uses VLM semantic consequence reasoning to \textit{infer per-object traversal risk conditioned on the robot's morphology and locomotion modality, without task-specific training}.
    \item A \textbf{visuosemantic caching mechanism} that uses CLIP embeddings and a vector store to detect semantically recurrent scenes, \textit{reusing prior risk assessments and significantly reducing online LLM query latency.}
    \item A \textbf{VLM-based trajectory reasoning module} that visually evaluates multi-proposal paths overlaid on the RGB image, \textit{selecting the safest trajectory given behavioral constraints and robot capabilities.}
\end{enumerate}

\section{Related Work}

\subsection{Traversability Estimation}
Traditional traversability estimation methods rely on geometric cues like elevation, slope estimation and surface roughness \cite{wermelinger2016navigation,chilian2009}. While effective in structured environments, these methods do not generalize well needing environment specific tuning. They do not account for semantic or contextual factors which influence navigation behavior. To overcome these limitations, supervised learning based methods have been introduced that integrate semantic information about the environment \cite{pmlr-v164-shaban22a,Maturana2017RealTimeSM}. These methods capture terrain characteristics beyond geometry. However, they require extensive data collection and expert annotations to achieve state of-
the-art-performance.

More recently, self supervised methods have reduced the reliance on manual labels \cite{wellhausen2019,gasparino2022wayfast,gasparino2024wayfaster,frey23fast,gummadi2024fed}. WayFaster \cite{gasparino2024wayfaster} learns traversability prediction by interacting with the environment to measure traction coefficients which are then projected onto camera images. Similarly, WVN \cite{frey23fast} estimates traversability online using DINO features and discrepancies between commanded and actual robot velocities.  But these methods require the robot to interact with the environment to generate labels which could be dangerous in hazardous environments. Weakly supervised methods allow selective manual annotations enabling labeling in untraversable area \cite{wrizz2024,schreiber_you_2025}. W-RIZZ \cite{wrizz2024} formulates relative traversability learning while CHUNGUS \cite{schreiber_you_2025} combines novelty detection with selective human annotation to improve generalization to unseen areas.

\subsection{Vision-Language Model assisted Navigation}
Advances in vision-language models (VLMs) have enabled the integration of semantic reasoning into navigation pipelines \cite{ahn2022can,shah2023lm,dorbala2022clip,huang_visual_2023}. Several approaches augment classical planners with cost-maps derived from VLM commonsense reasoning \cite{du_vl-nav_2025,weerakoon2025behav,yokoyama2024vlfm,gummadi2025zest}. For instance, BehAV \cite{weerakoon2025behav} encodes socially aware constraints into a behavioral cost-map. Similarly, VLFM \cite{yokoyama2024vlfm} generates a language-guided value map that drives exploration towards instruction consistent areas.
Other methods like ConVOI \cite{sathyamoorthy2024convoi} use VLMs to generate trajectories by reasoning over scene context, behavior guidelines and goal.
On the other hand, at the control layer, VLM-Social-Nav \cite{song2024vlm} and Hey Robot \cite{martinez2025hey} embed VLM based scoring functions within low-level motion planners.

\subsection{Vision-Language-Action Models for Navigation}
Vision-Language-Action models (VLA) represent a growing paradigm in navigation, aiming to learn end-to-end policies that directly map multimodal inputs to low-level control actions. OmniVLA \cite{hirose_omnivla_2025} extends this paradigm by training a VLA backbone which allows goals in multiple modalities to improve robustness across navigation tasks. VAMOS \cite{castro_vamos_2025} adopts a hierarchical design that uses the VLA to generate candidate action proposals which are then refined via reinforcement learning. NaVILA \cite{cheng2025navila} predicts mid-level actions from the VLA which are then fed as input to an RL policy.

\section{Method}
In this section, we present CATNAV, a framework featuring a novel method for generating traversability maps based on risk scores evaluated for each object in the robot's surrounding environment. This approach leverages the common-sense knowledge of Large Language Models (LLMs) to utilize open-vocabulary segmentation, \textit{while specifically addressing the latency of frequent online queries through a visuosemantic caching mechanism}. The resulting traversability map is then integrated with sampling-based trajectory optimization to generate proposals paths, from which the best candidate is selected using LLM reasoning to ensure safe travel.
\begin{figure*}[t!]
\centering
\includegraphics[width=0.95\textwidth]{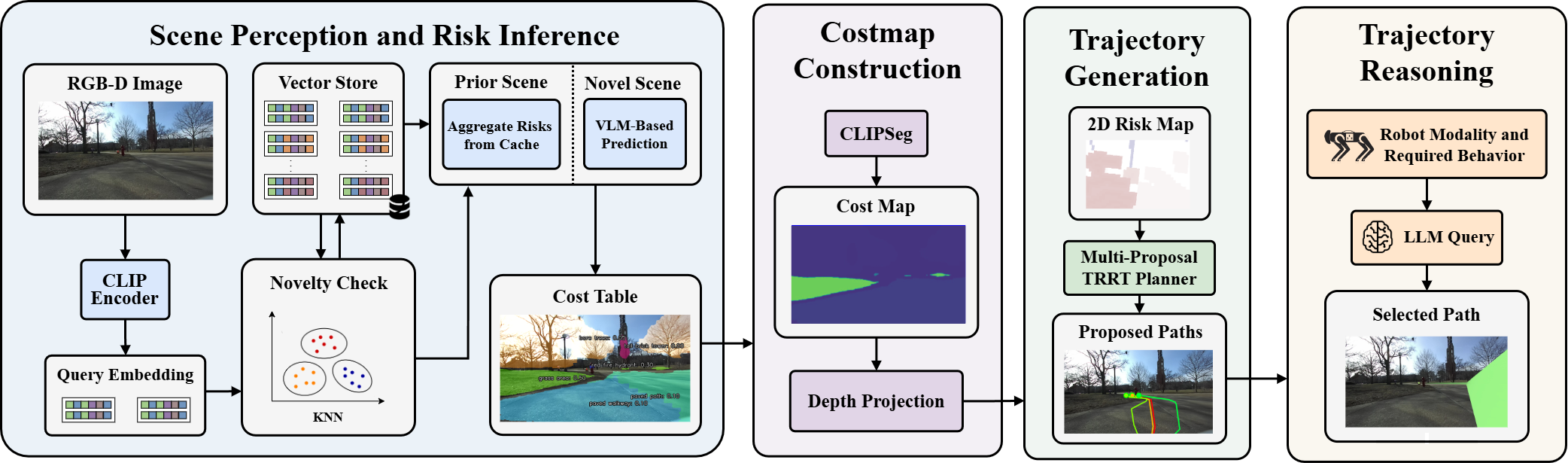}
\caption{Overview of the CATNAV pipeline framework for robot navigation. The system integrates real-time scene perception with risk inference, utilizing a Novelty Check module to determine if a scene requires a new VLM-based costmap prediction or can be processed via cached risks. Following Costmap Construction and Multi-Proposal TRRT Planning, the Trajectory Reasoning module employs a Large Language Model (LLM) to evaluate the proposed paths against specific robot modalities and required behaviors, ultimately selecting the single optimal trajectory for execution.}
\label{fig:catnav_overview}
\end{figure*}

\subsection{Scene Perception and Risk Inference}

\subsubsection{Cost Estimation}
\label{subsec:cost_estimation}

The initial stage of our framework involves extracting environmental semantics from the robot's visual input. Given an RGB observation $I \in \mathbb{R}^{H \times W \times 3}$, we leverage the zero-shot reasoning of a multimodal LLM (e.g., Gemini 3.0 Flash) to evaluate the traversability of the surrounding terrain.

To ground this assessment in the robot's physical reality, we provide a structured behavioral prompt that conditions the model on the robot's specific morphology. The LLM identifies objects, evaluates navigation risks, and calculates a semantic ``curiosity" score, mapping visual inputs to a structured cost table $\mathcal{T} = \{(c_k, r_k)\}_{k=1}^{K}$. Here, $c_k$ is a natural-language class label, $r_k \in [0, 1]$ is the traversal cost, and $K$ is the total number of identified classes, using the following prompt:

\begin{tcolorbox}[colback=gray!5,colframe=black]
    \small
    ``Identify all the objects seen (simple 1 word names) in the image and risk those objects may have while navigating if the robot collides with them or is hard to traverse. Also mention overall description of the current scene. (40 character limit). Assign a risk cost in range of 0.0 to 1.0, where 0.0 is no risk and 1.0 is the highest risk. Assign a curiosity score in range of 0.0 to 1.0 for the objects, where objects with similar place of use are more curious. e.g.\ a table and desk are closer to each other in meaning with higher score than car and desk. If the object is risky to go over give that object a high cost, otherwise give the ground a low cost. Keep previously seen items in list. The robot is \texttt{<modality>}''
\end{tcolorbox}
\noindent where the \texttt{<modality>} is replaced with the robot’s locomotion modality and physical dimensions (e.g., ``a wheeled robot, 0.3\,m ground clearance'').

\subsubsection{Novelty Detection \& Risk-Score Caching}
\label{subsec:novelty}

To minimize redundant LLM queries and reduce computational latency, we implement a caching mechanism that triggers a new risk-score analysis only when the visual context significantly diverges from previously processed scenes.

 Each incoming RGB frame $I_i$ is projected into a 768-dimensional latent space vector $\mathbf{z}_q$ using a CLIP image encoder. This query embedding is then compared against a history of $N$ stored embeddings stored in the vector store $\mathcal{S} =  \{\mathbf{z}_i\}_{i=1}^N$. We identify the $k$-nearest neighbors $\mathcal{N}_k$ by minimizing the Euclidean distance:

\begin{equation}
\mathcal{N}_k = \underset{k}{\arg\min} \|\mathbf{z}_q - \mathbf{z}_i\|_2, \quad i \in \{ 1, \dots, N\}.
\end{equation}

Since CLIP embeddings are unit-normalized, their pairwise distances lie in $[0, 2]$. In addition, we define the minimum distance \(d_{\min}\) as the distance between the query embedding and its nearest neighbor set. This value is compared against a novelty threshold $\gamma$ to determine if a new LLM query is required.

If $d_{min} \leq \gamma$, the frame is considered visually redundant. In this case, rather than querying the LLM, a cached cost table is constructed by averaging the risk scores across $\mathcal{N}_k$. Otherwise, the LLM is queried with the current frame to produce a new cost table, which is appended to the vector store $\mathcal{S}$ alongside its associated query embedding.

The aggregated risk is obtained via Eq.~\eqref{eq:cached_cost}, evaluated separately for each class present in any of the cached neighbor tables, yielding the final aggregated cost table. This caching strategy significantly reduces online inference latency during repeated traversals of visually similar terrain.

\begin{equation}
    \bar{r}_c = \frac{1}{|C|} \sum_{i \in C} r_c^{(i)},\quad C = \{ i \in \mathcal{N}_k \mid c \in \mathcal{T}_i \}\, .
    \label{eq:cached_cost}
\end{equation}

\subsection{Costmap Construction}
\label{subsec:costmap}
\subsubsection{Open-Vocabulary Segmentation and Cost Projection}
\label{subsubsec:segmentation}

Building on the perception output described in the previous section, we project the cost table into a costmap using the open-vocabulary capabilities of CLIPSeg \cite{luddecke2022image}. This process generates a dense, per-pixel cost segmentation image by grounding the semantic labels into the spatial domain.

Using the cost table $\mathcal{T}$, we tokenize all class labels $c_k$ and supplement them with pre-defined background prompts representing non-traversable or semantically null regions (e.g., "background," "sky," or "nothing"). CLIPSeg decodes the image embeddings conditioned on these tokens, where each prompt generates a corresponding logit map $l_k \in \mathbb{R}^{h \times w}$.

Since the decoded logit maps are produced at a lower resolution than the input, we apply bilinear interpolation to upsample them to the original image dimensions ($H \times W$). Finally, a softmax operation is applied across all channels to obtain a normalized probability distribution $p_k(u, v)$, representing the probability that pixel $(u, v)$ belongs to a specific terrain class $c_k$. The background class is used for normalization but is then discarded in subsequent steps.

With this, we can compute the per‑pixel costmap using Eq.~(\ref{eq:map_gen}), by assigning to each pixel the risk associated with its highest‑scoring class, provided the corresponding confidence exceeds a given threshold. It is worth highlighting that unassigned pixels are assigned zero cost, and the risk values $r_k$ are derived either from a fresh LLM query or the aggregated risk.
\begin{align}
    M(u,v) &=
    \begin{cases}
    r_{k_c} & \text{if } p_{k_c}(u,v) > \delta, \\
    0 & \text{otherwise},
    \end{cases}, \nonumber\\
    k_c &= \arg\max_k \, p_k(u,v) \cdot r_k,
    \label{eq:map_gen}
\end{align}
where $M$ is the resulting costmap, and $\delta$ is the confidence threshold used to filter out uncertain or weak segmentations.

\subsubsection{Risk-Scored Point Cloud Generation}
\label{subsubsec:pointcloud}

Given the pixel-wise costmap $\mathcal{M}$ in the image domain, we project the semantic risks into 3D space using synchronized depth information. Each pixel with a valid depth and a cost is back-projected into a 3D point using the camera intrinsic matrix. Each resulting point in the coordinate frame carries the specific traversal risk derived from the semantic costmap.

To maintain real-time performance on embedded hardware and prevent redundant point density, we employ two levels of data reduction. First, we apply strided pixel sampling during the initial back-projection; second, we process the resulting point cloud through a voxel grid filter. This down-sampling ensures that each voxel retains only a single representative point.

\subsubsection{2D Occupancy Costmap}
\label{subsubsec:2dcostmap}

To avoid complex 3D motion planning, we project the semantic risks into a 2D occupancy grid $G(x,y)$ by collapsing the vertical layers into a top‑down representation. For each grid cell $(x, y)$, we retain the maximum risk score observed across all corresponding heights, ensuring a conservative safety margin for navigation.

To handle occlusions and map updates, we apply Bresenham’s ray‑tracing algorithm starting from the sensor origin. This process marks the cells along the rays between the robot and the observed points as traversable, while regions beyond the sensor range remain classified as unknown.

\subsection{Goal Specification}
\label{subsec:goal}

CATNAV supports multiple goal modalities. In the vision-based mode, the LLM is queried to identify a normalized image-space point corresponding to the goal (e.g., a cone or doorway). This point is combined with the aligned depth map to recover a 3D pose subgoal in the robot's frame. Alternatively, the system accepts GPS waypoints fused with local odometry to produce metric pose goals, enabling long-range outdoor navigation without visual goal detection. Both modalities feed into the same TRRT planning pipeline.

\subsection{Trajectory Generation}
\label{subsec:trajgen}
\subsubsection{TRRT-Based Path Planning}
\label{subsubsec:trrt}

We employ a Transition‑based RRT (TRRT)~\cite{jaillet_transition-based_2008} motion‑planning method to compute cost‑aware paths over \(G\). This approach extends the basic RRT sampling strategy by accepting uphill cost transitions with a Boltzmann‑like probability, allowing us to tune parameters that balance exploration and exploitation. With this, the planner generates paths that not only minimize distance but also incorporate our costmap into the formulation.

 In addition, when the goal lies beyond the sensor horizon, planning is redirected to the nearest safe frontier cell along the robot‑to‑goal ray. The resulting path is then refined through shortcutting, pruning, and resampling to facilitate downstream MPPI tracking.

\subsubsection{Multi-Proposal Generation}
\label{subsubsec:multiproposal}
To promote spatial diversity, the planner generates four candidate paths from the TRRT. The first, generated with the standard settings, serves as the center path. By applying an offset $b$ to this center path, we obtain two additional variants oriented to the right and left. Offset positions that fall in unsafe cells are progressively shrunk until they become feasible. A fourth, “risky” path relaxes the cost ceiling to allow traversal through higher‑cost regions when the conservative proposals are blocked or excessively long. The four labeled proposals are then published for evaluation by the trajectory‑reasoning module.
\subsection{Trajectory Reasoning}
\label{subsec:trajreason}
The candidate path proposals are evaluated by a secondary LLM query (Gemini 3.0 Flash) that performs visual trajectory reasoning in the image domain. Each path is projected onto the RGB frame using a distinct color, and the resulting annotated image is provided to the LLM along with the following behavioral prompt:

\begin{tcolorbox}[colback=gray!5,colframe=black]
\small
``Check which color path line is the best to take for a \texttt{<modality>} robot with the behavior requirement \texttt{<behavior>}. Respond in two lines: Reason: and Color: \# or `none' if no goal point or viable safe path  visible.''
\end{tcolorbox}

\noindent The \texttt{<modality>} field corresponds to the robot's physical constraints as defined in Section~\ref{subsec:cost_estimation}, while \texttt{<behavior>} encodes mission-specific preferences such as ``stay left'', ``prefer the center'', or ``avoid crops''. This architecture decouples physical traversability from high-level behavioral objectives, allowing the same pipeline to be adapted to different platforms and missions.

To ensure temporal consistency, the LLM uses near-zero temperature and a sliding history of the previous exchanges. The path selected, identified by its color assignment, is then passed to a Nonlinear Model Predictive Controller (NMPC) for tracking. To maintain planning frequency, these queries are executed asynchronously with rate-limiting to prevent computational bottlenecks.

\section{Experiments}
In this section, we present the implementation of CATNAV in
the real world and analyze the experimental results.

\subsection{Implementation}
We deployed CATNAV on a Unitree Go1, a compact quadruped robot. The Go1's onboard hardware includes four legs, an IMU, and a stereo camera array. To support CATNAV's sensing and navigation requirements, we augmented the platform with a Global Navigation Satellite System (GNSS) receiver and a Stereolabs ZED 2i camera, which provides stereo depth perception and Visual-Inertial Odometry (VIO). All onboard computation is handled by an NVIDIA Jetson Orin. To enable real-time LLM API queries, we integrated a 4G/LTE GSM router for reliable internet connectivity in outdoor environments.

\subsection{Test Scenarios}
We evaluate across five tasks spanning outdoor and indoor environments. In the outdoor tasks (Tasks~1--4), the robot navigates a footpath toward a goal cone placed 20\,m away, with behavioral constraints varying per task (e.g., stay right, stay centered, avoid benches). In the dynamic scene task (Task~3), a human crosses the robot's path at the halfway point of a 17\,m course. Task~5 is an indoor scenario where the robot must reach a door while avoiding walking over paper placed on the floor. Each task was evaluated over N=10 trials per method. All trials per task were conducted in a single session under consistent environmental conditions.

\begin{figure}[b]
\centering
\setlength{\tabcolsep}{1pt}
\begin{tabular}{cc}
\includegraphics[width=0.47\columnwidth,height=2.35cm,keepaspectratio,clip,trim=0 0 0 0]{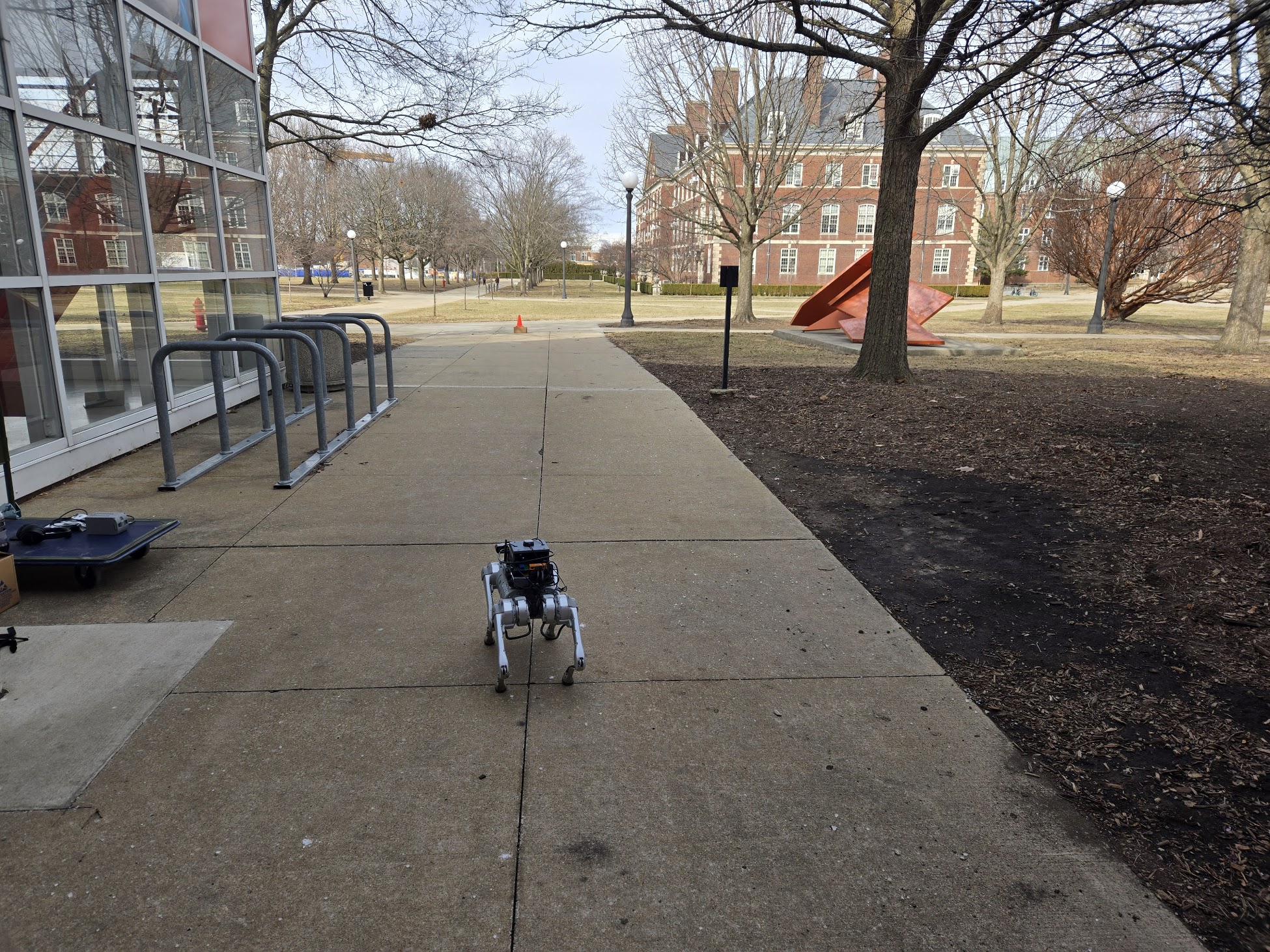} &
\includegraphics[width=0.48\columnwidth,height=2.5cm,keepaspectratio,clip,trim=0 0 0 0]{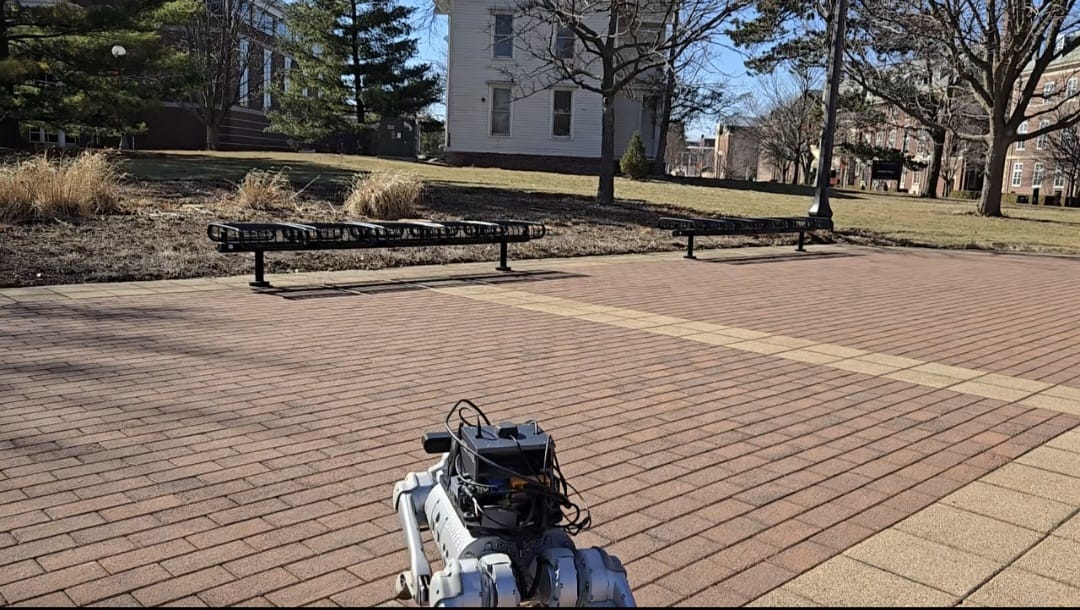} \\
{\small (a) Outdoor footpath} & {\small (b) Outdoor with obstacles} \\[2pt]
\includegraphics[width=0.48\columnwidth,height=2.33cm,keepaspectratio,clip,trim=0 0 0 0]{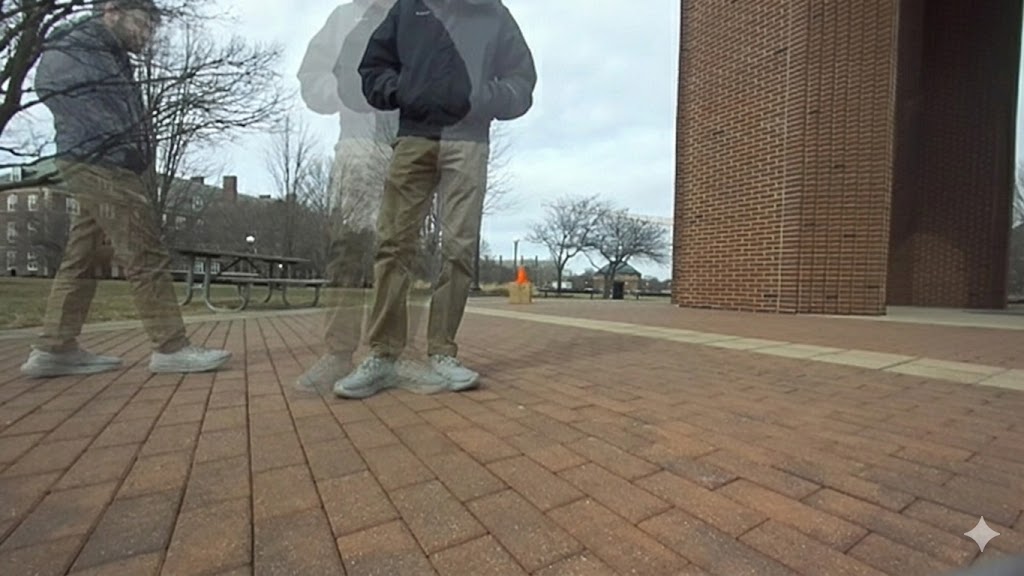} &
\includegraphics[width=0.48\columnwidth,height=2.5cm,keepaspectratio,clip,trim=0 0 0 0]{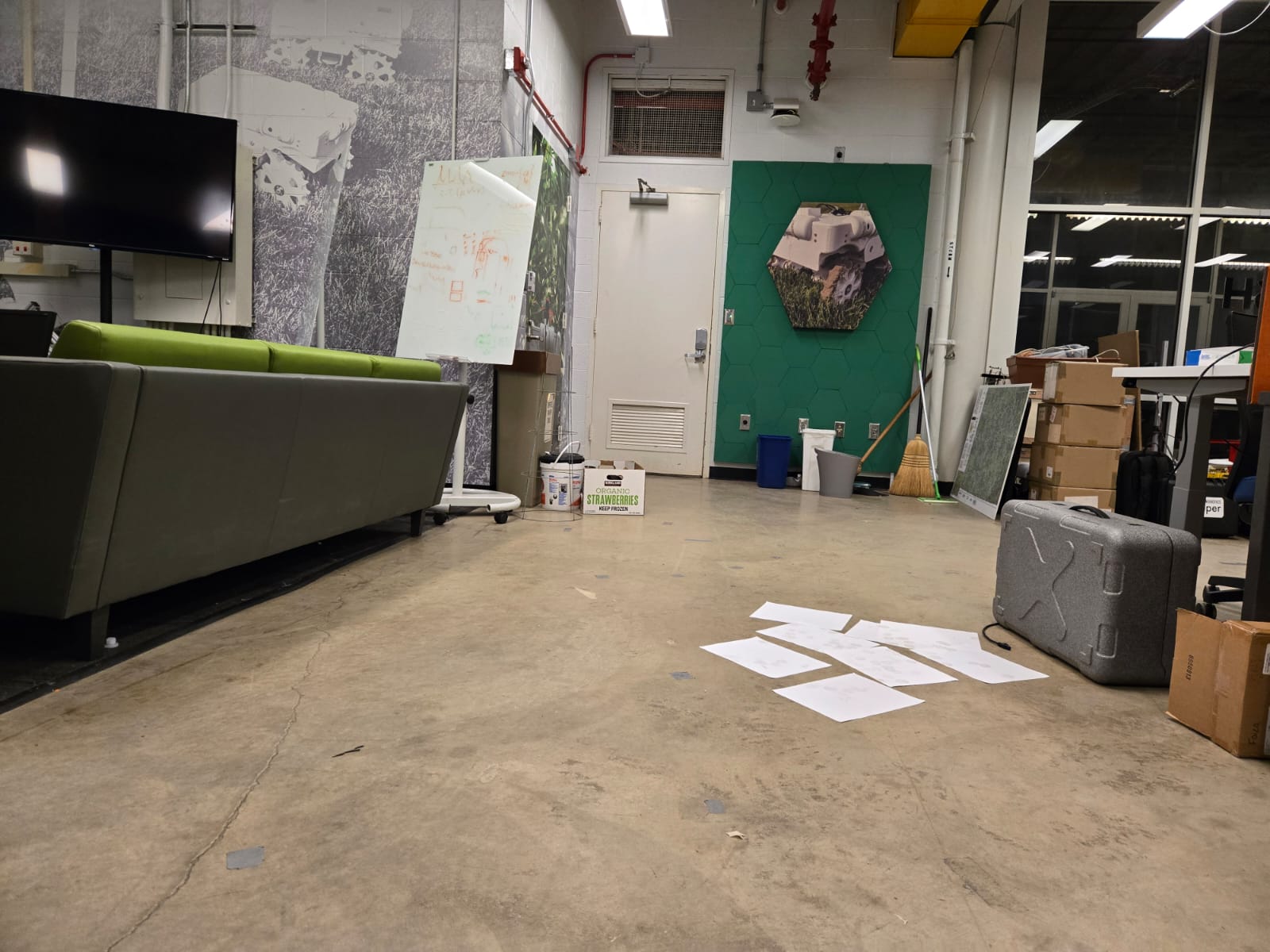} \\
{\small (c) Dynamic obstacle} & {\small (d) Indoor environment} \\
\end{tabular}
\caption{Scenarios samples: (a)~outdoor footpath navigation (Tasks~1--2), (b)~outdoor navigation with obstacles (Task~4), (c)~dynamic human crossing (Task~3), (d)~indoor paper avoidance (Task~5).}
\label{fig:test_scenarios}
\end{figure}

\subsection{Caching Mechanism Ablation}

Before evaluating the complete system, we conduct a series of ablation studies to analyze key parameters of the visuosemantic caching mechanism. Specifically, we examine the number of nearest neighbors $k$ used for cost-table aggregation, and the novelty threshold $\gamma$, which determines when a new LLM query is triggered. Table~\ref{tab:vlm_query_rate} summarizes the configurations for the three test settings, where the total queries represent the sum of high-level scene understanding and specific path selection requests over approximately two hours of driving per test. For reference, we also include a fixed pooling rate of 2Hz, as employed by ConVOI.

Analyzing the results in Table~\ref{tab:vlm_query_rate}, we observe that setting $\gamma=0.1$ with $k=5$ effectively disables caching, since nearly every frame is considered novel. This leads to a 10x increase in query rate compared to the first setting. In contrast, Test 3, which uses $\gamma=0.55$ and $k=5$, achieves the best polling performance—reducing VLM queries by 85.7\% and increasing cache utilization by 86.5\% relative to Test 1. These results suggest that aggregating risk estimates across multiple similar scenes leads to more stable cached representations, enabling the system to rely more heavily on cached information without compromising costmap quality.

\begin{figure}[b]
\centering
\includegraphics[width=\columnwidth]{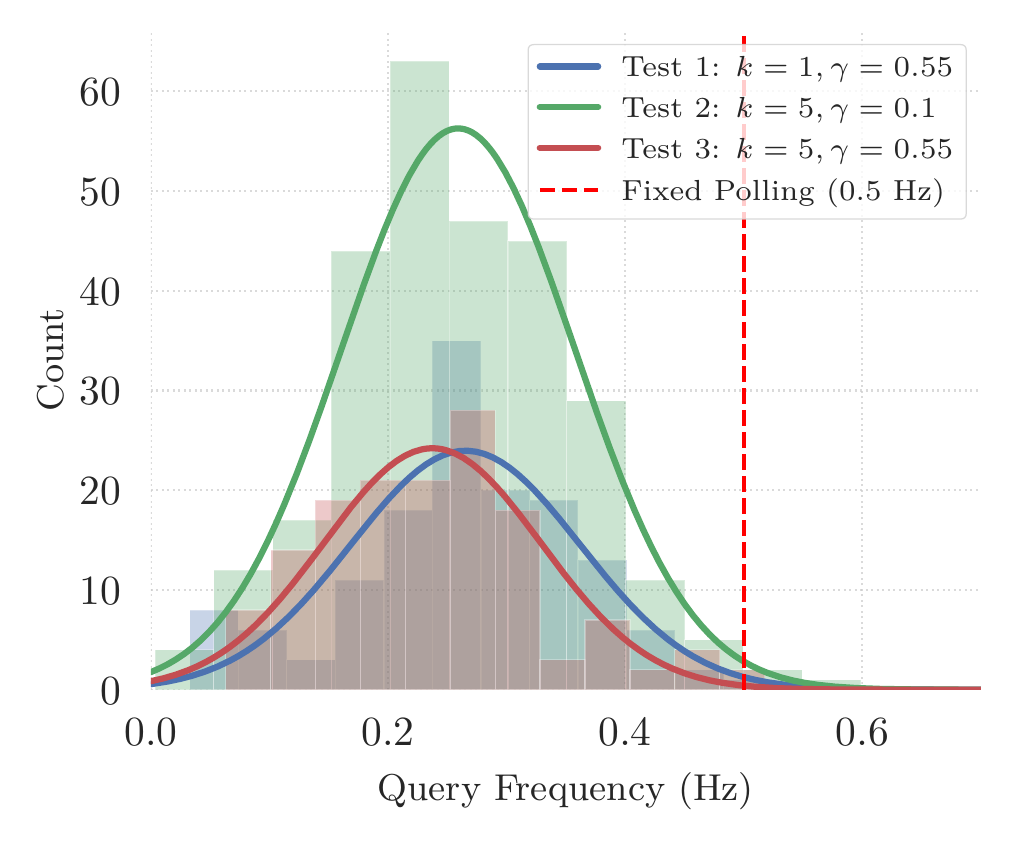}
\caption{Distribution of VLM query frequencies across different caching configurations. The histograms represent the raw query counts, overlaid with Gaussian approximations scaled to the total number of samples and bin width for each test. A reference fixed frequency is plotted based on the fixed polling baseline.}
\label{fig:hist}
\end{figure}

Furthermore, Fig.~\ref{fig:hist} illustrates the distribution of query frequencies along with a Gaussian approximation. This allows us to visualize the average query rate and frequency distribution for each configuration, offering clearer insight into how the third setting of our proposed caching mechanism achieves the observed reduction in VLM calls.

\begin{table}[t]
\centering
\caption{VLM Query Rate Comparison Across Configurations.}
\renewcommand{\arraystretch}{1.1}
\setlength{\tabcolsep}{10pt}
\begin{tabular}{l c c c}
\toprule
& \textbf{Test 1} & \textbf{Test 2} & \textbf{Test 3} \\
\midrule
Neighbors ($k$)         & 1               & 5               & 5               \\
Novelty Threshold ($\gamma$)    & 0.55            & 0.1             & 0.55            \\
\midrule
$\downarrow$ Scene Queries           & 16              & 125             & 9               \\
$\downarrow$ Path Selector Calls     & 156             & 295             & 157             \\
$\downarrow$ Total VLM Queries       & 172             & 420             & 166             \\
\midrule
$\uparrow$ Cache Rate (C/s)          & 0.044           & 0.104           & 0.194           \\
$\downarrow$ Query Rate (Q/s)        & 0.004           & 0.042           & 0.006           \\
$\downarrow$ Avg Latency (s)         & 3.937           & 17.35           & 3.717           \\
\bottomrule
\end{tabular}
\label{tab:vlm_query_rate}
\end{table}

\subsection{Scene Perception}

\begin{figure}[t]
\centering
\includegraphics[width=\columnwidth]{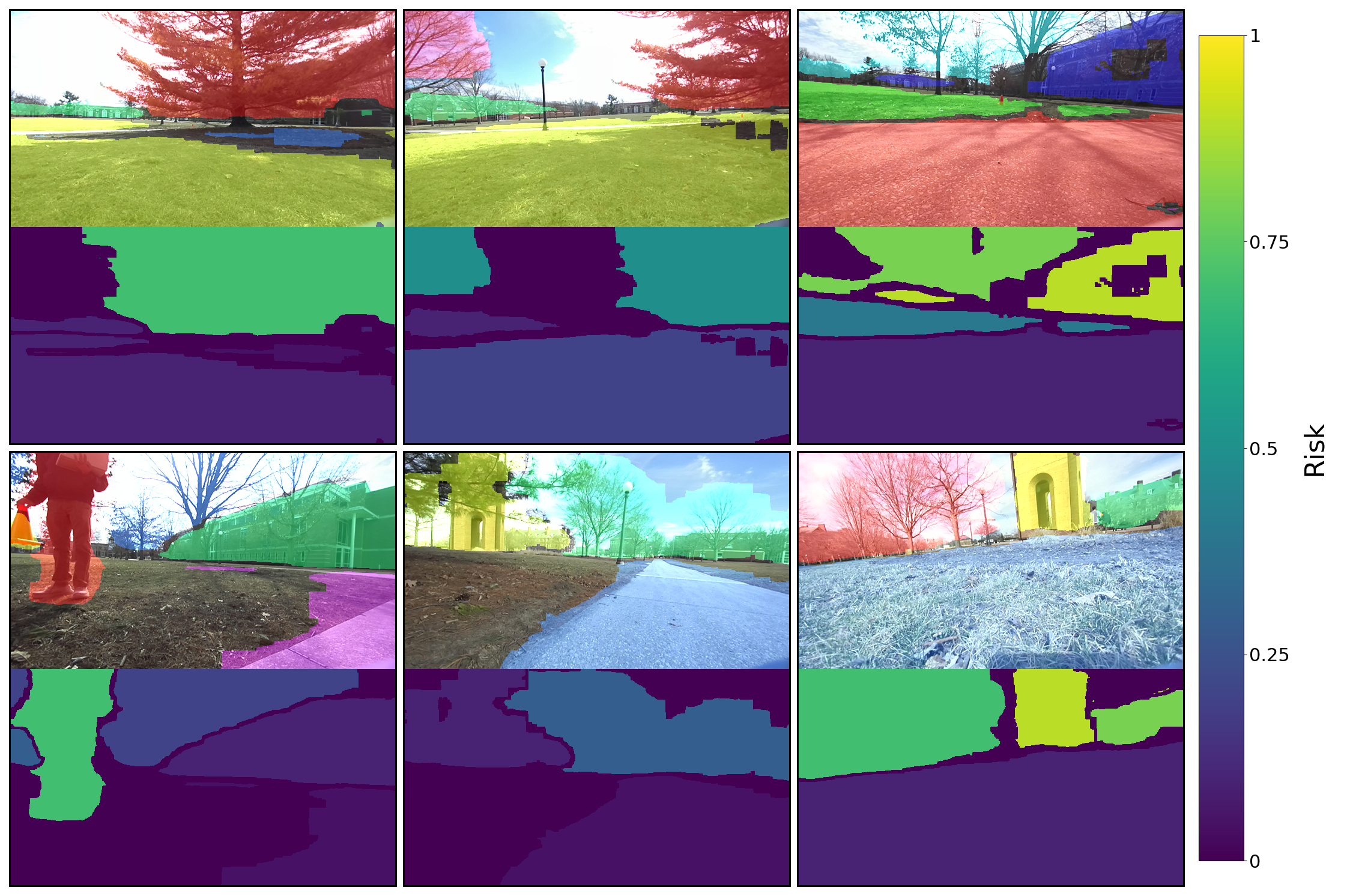}
\caption{Qualitative results of the costmap segmentation using CATNAV's visuosemantic cache. The cost table is aggregated from the $k$-nearest neighbors of the current scene's CLIP embedding. If no neighbors are found within the novelty threshold $\gamma$, the scene is classified as "Novel," queried via the LLM, and appended to the vector store.}
\label{fig:qualitative}
\vspace{-5pt}
\end{figure}

To visualize the performance of the scene perception module, Fig.~\ref{fig:qualitative} displays the generated segmentation maps alongside the corresponding costmaps, produced by projecting class-specific costs into the image domain. As shown in the images, the system dynamically assigns cost values based on traversability: benign terrain (e.g., pavement) receives a low cost of 0.1, partially traversable regions (e.g., grass) are assigned a moderate cost of 0.3, and significant hazards (e.g., pedestrians) are marked with a high cost of 0.7.

\subsection{System Evaluation}

Based on the ablation results, we select the optimal configuration for subsequent experiments to comprehensively evaluate our system. Specifically, we employ the following metrics:

\begin{itemize}
    \item Goal Reaching (\%). The percentage of trials in which the robot reaches within a fixed distance threshold of the goal, measuring overall navigation reliability.
    \item Distance to Goal (m). The Euclidean distance between the robot's final position and the goal at trial termination. Lower values indicate more precise navigation, even in cases of partial success.
    \item Collision Rate (\%). The percentage of trials in which the robot makes physical contact with an obstacle, reflecting trajectory safety and the costmap's effectiveness at encoding obstacle risk.
    \item Behavioral Constraint Violations (\%). The percentage of trials in which the robot violates a task-specific behavioral instruction (e.g., \textit{stay on the right side of the path''} or \textit{avoid walking over the paper''}), evaluating whether the method respects high-level navigational intent beyond basic obstacle avoidance.
\end{itemize}

We compare CATNAV against the vision-language-action model OmniVLA~\cite{hirose_omnivla_2025}  across all five predefined tasks, which encompass both indoor and outdoor environments with static and dynamic subjects. The results are summarized in Table~\ref{tab:analysis}. For the baseline, we employ the subgoal generation method described in Section~\ref{subsec:goal} to define the pose goal w.r.t the robot and the semantic behaviour requirement that we require.

\begin{table}[t]
\centering
\caption{Navigation performance of CATNAV vs. OmniVLA across five tasks (10 trials per task)}
\setlength{\tabcolsep}{3pt}
\begin{tabular}{l cccc}
\toprule
\textbf{Method} & \makecell{\textbf{Goal} \\ \textbf{Reaching(\%) $\uparrow$}} & \makecell{\textbf{Dist. to} \\ \textbf{Goal} (m) $\downarrow$} & \makecell{\textbf{Collisions} \\ \textbf{(\%) $\downarrow$}} & \makecell{\textbf{Behavior} \\ \textbf{Breaks(\%) $\downarrow$}} \\
\midrule
\rowcolor{blue!8} \multicolumn{5}{l}{\textbf{Task 1: Navigate to cone, stay on right side of path (Static)}} \\
OmniVLA         & 60    & 2.7    & 20    & 40    \\
CATNAV (Ours)   & 70    & 2.8    & 10    & 30    \\
\midrule
\rowcolor{blue!8} \multicolumn{5}{l}{\textbf{Task 2: Navigate to cone, stay in center of path (Static)}} \\
OmniVLA         & 50    & 2.3    & --    & 20    \\
CATNAV (Ours)   & 70    & 2.25    & --    & 20    \\
\midrule
\rowcolor{blue!8} \multicolumn{5}{l}{\textbf{Task 3: Navigate to cone, (Human crosses path, Dynamic)}} \\
OmniVLA         & 60    & 0.4    & 20    & --    \\
CATNAV (Ours)   & 70    & 0.3    & 20    & --    \\
\midrule
\rowcolor{blue!8} \multicolumn{5}{l}{\textbf{Task 4: Navigate to house (Avoid Collisions with Benches, Static)}} \\
OmniVLA         & 30    & 0.8    & 20    & --    \\
CATNAV (Ours)   & 40    & 0.7    & 30    & --    \\
\midrule
\rowcolor{blue!8} \multicolumn{5}{l}{\textbf{Task 5: Go to door, avoid walking over the paper on the floor (Static)}} \\
OmniVLA         & 90    & 0.1    & 0    & 70    \\
CATNAV (Ours)   & 90    & 0.1    & 0    & 30    \\
\bottomrule
\end{tabular}
\label{tab:analysis}
\vspace{-0.5em}
\end{table}

\subsubsection{Zero-Shot Generalization.}
CATNAV achieves a higher goal-reaching rate on four of five tasks, averaging 68\% compared to 58\% for OmniVLA, with the largest gains on Tasks~2 and~4 (+20\% and +10\% respectively).
For Tasks~1, we found that while OmniVLA was closer to goal object on average, it often parked on the side of the goal object instead in front of it. Furthermore, the baseline had more behaviour breaks such as crossing into the grass or colliding with obstacles. We theorize the cloudy weather during tests was out of distribution, which affected performance, whereas CATNAV was robust to the weather.
Because traversal risk is inferred at query time from the robot's morphology description rather than learned from a fixed training distribution, the same pipeline generalizes across indoor and outdoor environments without retraining. In the dynamic scene (Task~3), CATNAV achieves a 10\% higher goal-reaching rate while matching OmniVLA on collisions, indicating that the VLM-based costmap can identify and react to dynamic agents without specialized training.

\subsubsection{Behavioral Constraint Adherence.} The trajectory reasoning module explicitly conditions path selection on behavioral instructions, yielding a 33\% reduction in constraint violations across tasks where behavioral goals are specified. On Task~5, both methods reach the goal equally (90\%), yet CATNAV reduces behavior breaks from 70\% to 30\% by visually grounding the instruction ``avoid the paper'' during path selection. Similarly, on Task~1 CATNAV halves the collision rate (10\% vs.\ 20\%) while also lowering behavior breaks (30\% vs.\ 40\%). Task~4 reveals a current limitation: although goal reaching improves (40\% vs.\ 30\%), the collision rate increases (30\% vs.\ 20\%), suggesting the costmap can underestimate risk for narrow obstacles at longer ranges.

\subsubsection{Caching Efficiency and Robustness.} The visuosemantic caching mechanism (Table~\ref{tab:vlm_query_rate}) reduces VLM scene queries by 85.7\% while increasing cache utilization by 86.5\%, enabling costmap updates at up to 10\,Hz from cached risk scores during visually stable traversals. This allows the robot to maintain velocities up to 0.75\,m/s, slowing only when a high-risk dynamic object triggers replanning. The k-NN cost aggregation additionally smooths outlier risk estimates across similar cached scenes, reducing downstream sensitivity to individual LLM mispredictions.

\begin{figure}[t]
\centering
\setlength{\tabcolsep}{1pt}
\begin{tabular}{cc}
\includegraphics[width=0.44\columnwidth]{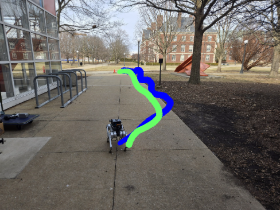} &
\includegraphics[width=0.44\columnwidth]{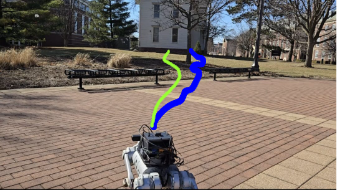} \\
{\small (a) Tasks 1--2: Footpath navigation} & {\small (b) Task 4: Bench avoidance} \\[2pt]
\includegraphics[width=0.44\columnwidth]{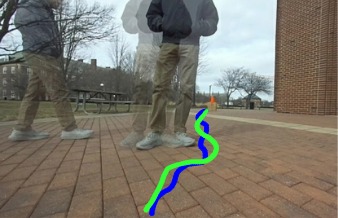} &
\includegraphics[width=0.44\columnwidth]{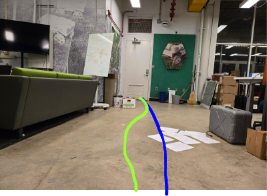} \\
{\small (c) Task 3: Dynamic obstacle} & {\small (d) Task 5: Paper avoidance} \\
\end{tabular}
\caption{Trajectory visualizations of CATNAV across all tasks. Colored lines denote candidate and selected trajectories overlaid on the robot's RGB view., with \textcolor{blue_traj}{\textbf{---}} representing the baseline and \textcolor{green_traj}{\textbf{---}} representing \textsc{CATNAV}.}
\label{fig:collage_trajs}
\end{figure}

Finally, Fig.~\ref{fig:collage_trajs} illustrates the best trajectories generated by our proposed method (green) versus the baseline (blue) across all five tasks. These visualizations substantiate our previous findings regarding zero-shot navigation in both static and dynamic environments. Beyond simple path completion, the trajectories demonstrate superior behavioral adherence.

For instance, in the indoor scenario, while the baseline could reach the goal, it failed to avoid obstacles like papers on the floor. Similarly, in the footpath task, the baseline was unable to track the designated path while simultaneously avoiding the grass.

\section{Conclusion}
We presented CATNAV, a framework for zero-shot, embodiment-aware robot navigation that leverages VLM semantic consequence reasoning for costmap generation and trajectory selection. By introducing a visuosemantic caching mechanism based on $k$-nearest neighbor retrieval over CLIP embeddings, CATNAV reduces online LLM costmap queries by 85.7\% without degrading navigation performance. A second VLM query enables visual trajectory reasoning that respects behavioral constraints specified through natural language prompts.

Experiments on a quadruped robot across 5 indoor and outdoor tasks demonstrate that CATNAV achieves a 10\% point higher average goal-reaching rate and 33\% fewer behavioral constraint violations compared to a state-of-the-art VLA baseline, while generalizing to new environments and tasks through prompt reconfiguration alone.

A current limitation is the reliance on internet connectivity for cloud-based LLM inference, which introduces latency and limits deployment in communication-denied environments. Future work will explore on-device distillation of the VLM reasoning into lightweight models, as well as extending the framework to multi-robot settings and longer-horizon planning tasks.
\vspace{-5pt}

\bibliographystyle{IEEEtran}
\bibliography{bibtex}

\begin{thebibliography}{10}
\providecommand{\url}[1]{#1}
\csname url@samestyle\endcsname
\providecommand{\newblock}{\relax}
\providecommand{\bibinfo}[2]{#2}
\providecommand{\BIBentrySTDinterwordspacing}{\spaceskip=0pt\relax}
\providecommand{\BIBentryALTinterwordstretchfactor}{4}
\providecommand{\BIBentryALTinterwordspacing}{\spaceskip=\fontdimen2\font plus
\BIBentryALTinterwordstretchfactor\fontdimen3\font minus \fontdimen4\font\relax}
\providecommand{\BIBforeignlanguage}[2]{{%
\expandafter\ifx\csname l@#1\endcsname\relax
\typeout{** WARNING: IEEEtran.bst: No hyphenation pattern has been}%
\typeout{** loaded for the language `#1'. Using the pattern for}%
\typeout{** the default language instead.}%
\else
\language=\csname l@#1\endcsname
\fi
#2}}
\providecommand{\BIBdecl}{\relax}
\BIBdecl

\bibitem{gasparino2024wayfaster}
M.~V. Gasparino, A.~N. Sivakumar, and G.~Chowdhary, ``Wayfaster: a self-supervised traversability prediction for increased navigation awareness,'' in \emph{2024 IEEE International Conference on Robotics and Automation (ICRA)}.\hskip 1em plus 0.5em minus 0.4em\relax IEEE, 2024, pp. 8486--8492.

\bibitem{wrizz2024}
A.~Schreiber, A.~N. Sivakumar, P.~Du, M.~V. Gasparino, G.~Chowdhary, and K.~Driggs-Campbell, ``W-rizz: A weakly-supervised framework for relative traversability estimation in mobile robotics,'' \emph{IEEE Robotics and Automation Letters}, vol.~9, no.~6, pp. 5623--5630, 2024.

\bibitem{schreiber_you_2025}
A.~Schreiber and K.~Driggs-Campbell, ``Do {You} {Know} the {Way}? {Human}-in-the-{Loop} {Understanding} for {Fast} {Traversability} {Estimation} in {Mobile} {Robotics},'' \emph{IEEE Robotics and Automation Letters}, vol.~10, no.~6, pp. 5863--5870, Jun. 2025.

\bibitem{frey23fast}
J.~Frey, M.~Mattamala, N.~Chebrolu, C.~Cadena, M.~Fallon, and M.~Hutter, ``{Fast Traversability Estimation for Wild Visual Navigation},'' in \emph{Proceedings of Robotics: Science and Systems}, Daegu, Republic of Korea, July 2023.

\bibitem{elnoor_vi-lad_2025}
M.~Elnoor, K.~Weerakoon, G.~Seneviratne, J.~Liang, V.~Rajagopal, and D.~Manocha, ``Vi-{LAD}: {Vision}-{Language} {Attention} {Distillation} for {Socially}-{Aware} {Robot} {Navigation} in {Dynamic} {Environments},'' Mar. 2025.

\bibitem{google2025gemini2.5}
G.~Comanici, E.~Bieber, M.~Schaekermann, I.~Pasupat, N.~Sachdeva, I.~Dhillon, M.~Blistein, O.~Ram, D.~Zhang, E.~Rosen \emph{et~al.}, ``Gemini 2.5: Pushing the frontier with advanced reasoning, multimodality, long context, and next generation agentic capabilities,'' \emph{arXiv preprint arXiv:2507.06261}, 2025.

\bibitem{luddecke2022image}
T.~L{\"u}ddecke and A.~Ecker, ``Image segmentation using text and image prompts,'' in \emph{Proceedings of the IEEE/CVF conference on computer vision and pattern recognition}, 2022, pp. 7086--7096.

\bibitem{radford2021learning}
A.~Radford, J.~W. Kim, C.~Hallacy, A.~Ramesh, G.~Goh, S.~Agarwal, G.~Sastry, A.~Askell, P.~Mishkin, J.~Clark \emph{et~al.}, ``Learning transferable visual models from natural language supervision,'' in \emph{International conference on machine learning}.\hskip 1em plus 0.5em minus 0.4em\relax PmLR, 2021, pp. 8748--8763.

\bibitem{du_vl-nav_2025}
\BIBentryALTinterwordspacing
Y.~Du, T.~Fu, Z.~Chen, B.~Li, S.~Su, Z.~Zhao, and C.~Wang, ``{VL}-{Nav}: {Real}-time {Vision}-{Language} {Navigation} with {Spatial} {Reasoning},'' Mar. 2025, arXiv:2502.00931 [cs]. [Online]. Available: \url{http://arxiv.org/abs/2502.00931}
\BIBentrySTDinterwordspacing

\bibitem{yokoyama2024vlfm}
N.~Yokoyama, S.~Ha, D.~Batra, J.~Wang, and B.~Bucher, ``Vlfm: Vision-language frontier maps for zero-shot semantic navigation,'' in \emph{2024 IEEE International Conference on Robotics and Automation (ICRA)}.\hskip 1em plus 0.5em minus 0.4em\relax IEEE, 2024, pp. 42--48.

\bibitem{weerakoon2025behav}
K.~Weerakoon, M.~Elnoor, G.~Seneviratne, V.~Rajagopal, S.~H. Arul, J.~Liang, M.~K.~M. Jaffar, and D.~Manocha, ``Behav: Behavioral rule guided autonomy using vlms for robot navigation in outdoor scenes,'' in \emph{2025 IEEE International Conference on Robotics and Automation (ICRA)}.\hskip 1em plus 0.5em minus 0.4em\relax IEEE, 2025, pp. 7044--7051.

\bibitem{sathyamoorthy2024convoi}
A.~J. Sathyamoorthy, K.~Weerakoon, M.~Elnoor, A.~Zore, B.~Ichter, F.~Xia, J.~Tan, W.~Yu, and D.~Manocha, ``Convoi: Context-aware navigation using vision language models in outdoor and indoor environments,'' in \emph{2024 IEEE/RSJ International Conference on Intelligent Robots and Systems (IROS)}.\hskip 1em plus 0.5em minus 0.4em\relax IEEE, 2024, pp. 13\,837--13\,844.

\bibitem{song2024vlm}
D.~Song, J.~Liang, A.~Payandeh, A.~H. Raj, X.~Xiao, and D.~Manocha, ``Vlm-social-nav: Socially aware robot navigation through scoring using vision-language models,'' \emph{IEEE Robotics and Automation Letters}, vol.~10, no.~1, pp. 508--515, 2024.

\bibitem{martinez2025hey}
D.~Martinez-Baselga, O.~de~Groot, L.~Knoedler, J.~Alonso-Mora, L.~Riazuelo, and L.~Montano, ``Hey robot! personalizing robot navigation through model predictive control with a large language model,'' in \emph{2025 IEEE International Conference on Robotics and Automation (ICRA)}.\hskip 1em plus 0.5em minus 0.4em\relax IEEE, 2025, pp. 11\,002--11\,009.

\bibitem{castro_vamos_2025}
M.~G. Castro, S.~Rajagopal, D.~Gorbatov, M.~Schmittle, R.~Baijal, O.~Zhang, R.~Scalise, S.~Talia, E.~Romig, C.~d. Melo, B.~Boots, and A.~Gupta, ``{VAMOS}: {A} {Hierarchical} {Vision}-{Language}-{Action} {Model} for {Capability}-{Modulated} and {Steerable} {Navigation},'' Oct. 2025.

\bibitem{hirose_omnivla_2025}
N.~Hirose, C.~Glossop, D.~Shah, and S.~Levine, ``{OmniVLA}: {An} {Omni}-{Modal} {Vision}-{Language}-{Action} {Model} for {Robot} {Navigation},'' Sep. 2025.

\bibitem{wermelinger2016navigation}
M.~Wermelinger, P.~Fankhauser, R.~Diethelm, P.~Kr{\"u}si, R.~Siegwart, and M.~Hutter, ``Navigation planning for legged robots in challenging terrain,'' in \emph{2016 IEEE/RSJ International Conference on Intelligent Robots and Systems (IROS)}.\hskip 1em plus 0.5em minus 0.4em\relax IEEE, 2016, pp. 1184--1189.

\bibitem{chilian2009}
A.~Chilian and H.~Hirschmüller, ``Stereo camera based navigation of mobile robots on rough terrain,'' in \emph{2009 IEEE/RSJ International Conference on Intelligent Robots and Systems}, 2009, pp. 4571--4576.

\bibitem{pmlr-v164-shaban22a}
A.~Shaban, X.~Meng, J.~Lee, B.~Boots, and D.~Fox, ``Semantic terrain classification for off-road autonomous driving,'' in \emph{Proceedings of the 5th Conference on Robot Learning}, ser. Proceedings of Machine Learning Research, A.~Faust, D.~Hsu, and G.~Neumann, Eds., vol. 164.\hskip 1em plus 0.5em minus 0.4em\relax PMLR, 08--11 Nov 2022, pp. 619--629.

\bibitem{Maturana2017RealTimeSM}
D.~Maturana, P.-W. Chou, M.~Uenoyama, and S.~Scherer, ``Real-time semantic mapping for autonomous off-road navigation,'' in \emph{Field and Service Robotics: Results of the 11th International Conference}.\hskip 1em plus 0.5em minus 0.4em\relax Springer, 2017, pp. 335--350.

\bibitem{wellhausen2019}
L.~Wellhausen, A.~Dosovitskiy, R.~Ranftl, K.~Walas, C.~Cadena, and M.~Hutter, ``Where should i walk? predicting terrain properties from images via self-supervised learning,'' \emph{IEEE Robotics and Automation Letters}, vol.~4, no.~2, pp. 1509--1516, 2019.

\bibitem{gasparino2022wayfast}
M.~V. Gasparino, A.~N. Sivakumar, Y.~Liu, A.~E. Velasquez, V.~A. Higuti, J.~Rogers, H.~Tran, and G.~Chowdhary, ``Wayfast: Navigation with predictive traversability in the field,'' \emph{IEEE Robotics and Automation Letters}, vol.~7, no.~4, pp. 10\,651--10\,658, 2022.

\bibitem{gummadi2024fed}
S.~Gummadi, M.~V. Gasparino, D.~Vasisht, and G.~Chowdhary, ``Fed-ec: Bandwidth-efficient clustering-based federated learning for autonomous visual robot navigation,'' \emph{IEEE Robotics and Automation Letters}, vol.~9, no.~12, pp. 11\,841--11\,848, 2024.

\bibitem{ahn2022can}
M.~Ahn, A.~Brohan, N.~Brown, Y.~Chebotar, O.~Cortes, B.~David, C.~Finn, C.~Fu, K.~Gopalakrishnan, K.~Hausman \emph{et~al.}, ``Do as i can, not as i say: Grounding language in robotic affordances,'' \emph{arXiv preprint arXiv:2204.01691}, 2022.

\bibitem{shah2023lm}
D.~Shah, B.~Osi{\'n}ski, S.~Levine \emph{et~al.}, ``Lm-nav: Robotic navigation with large pre-trained models of language, vision, and action,'' in \emph{Conference on robot learning}.\hskip 1em plus 0.5em minus 0.4em\relax PMLR, 2023, pp. 492--504.

\bibitem{dorbala2022clip}
V.~S. Dorbala, G.~Sigurdsson, R.~Piramuthu, J.~Thomason, and G.~S. Sukhatme, ``Clip-nav: Using clip for zero-shot vision-and-language navigation,'' \emph{arXiv preprint arXiv:2211.16649}, 2022.

\bibitem{huang_visual_2023}
\BIBentryALTinterwordspacing
C.~Huang, O.~Mees, A.~Zeng, and W.~Burgard, ``Visual {Language} {Maps} for {Robot} {Navigation},'' in \emph{2023 {IEEE} {International} {Conference} on {Robotics} and {Automation} ({ICRA})}.\hskip 1em plus 0.5em minus 0.4em\relax London, United Kingdom: IEEE, May 2023, pp. 10\,608--10\,615. [Online]. Available: \url{https://ieeexplore.ieee.org/document/10160969/}
\BIBentrySTDinterwordspacing

\bibitem{gummadi2025zest}
S.~Gummadi, M.~V. Gasparino, G.~Capezzuto, M.~Becker, and G.~Chowdhary, ``Zest: an llm-based zero-shot traversability navigation for unknown environments,'' \emph{arXiv preprint arXiv:2508.19131}, 2025.

\bibitem{cheng2025navila}
A.-C. Cheng, Y.~Ji, Z.~Yang, Z.~Gongye, X.~Zou, J.~Kautz, E.~B{\i}y{\i}k, H.~Yin, S.~Liu, and X.~Wang, ``Navila: Legged robot vision-language-action model for navigation,'' in \emph{RSS}, 2025.

\bibitem{jaillet_transition-based_2008}
L.~Jaillet, J.~Cortes, and T.~Simeon, ``Transition-based {RRT} for path planning in continuous cost spaces.''\hskip 1em plus 0.5em minus 0.4em\relax Nice: IEEE, Sep. 2008, pp. 2145--2150.

\end{thebibliography}
\end{document}